\documentclass[pdflatex,sn-mathphys-num]{sn-jnl}

\usepackage{graphicx}%
\usepackage{multirow}%
\usepackage{amsmath,amssymb,amsfonts}%
\usepackage{mathrsfs}%
\usepackage[title]{appendix}%
\usepackage{xcolor}%
\usepackage{textcomp}%
\usepackage{manyfoot}%
\usepackage{booktabs}%
\usepackage{algpseudocode}%
\usepackage{listings}%
\usepackage{tikz}
\usepackage{hyperref}
\usepackage{subfig}

\lstdefinestyle{pythonstyle}{
    language=Python,
    basicstyle=\ttfamily\footnotesize,
    keywordstyle=\bfseries\color{blue},
    stringstyle=\color{purple},
    commentstyle=\color{gray},
    numbers=none,
    numberstyle=\tiny\color{gray},
    stepnumber=1,
    numbersep=10pt,
    showstringspaces=false,
    frame=single,
    breaklines=true
}

\lstdefinestyle{conjecturestyle}{
    basicstyle=\ttfamily\footnotesize,
    keywordstyle=\bfseries\color{blue},
    commentstyle=\color{gray},
    showstringspaces=false,
    frame=single,
    breaklines=true,
    numbers=none,
    numberstyle=\tiny\color{gray},
}

\usepackage[ruled,vlined]{algorithm2e}

\begin{document}

\title[Article Title]{The \emph{Optimist}: Towards Fully Automated \\ Graph Theory Research}


\author*[1]{\fnm{Randy} \sur{Davila}}\email{rrd6@rice.edu}

\affil*[1]{\orgdiv{Department of Computational Applied Mathematics \& Operations Research}, \orgname{Rice University }, \orgaddress{\street{6100 Main Street}, \city{Houston}, \postcode{77005}, \state{TX}, \country{USA}}}


\abstract{This paper introduces the \emph{Optimist}, an autonomous system developed to advance automated conjecture generation in graph theory. Leveraging mixed-integer programming (MIP) and heuristic methods, the \emph{Optimist} generates conjectures that both rediscover established theorems and propose novel inequalities. Through a combination of memory-based computation and agent-like adaptability, the \emph{Optimist} iteratively refines its conjectures by integrating new data, enabling a feedback process with minimal human (\emph{or machine}) intervention. Initial experiments reveal the \emph{Optimist}’s potential to uncover foundational results in graph theory, as well as to produce conjectures of interest for future exploration. This work also outlines the \emph{Optimist}’s evolving integration with a counterpart agent, the \emph{Pessimist} (a human \emph{or machine} agent), to establish a dueling system that will drive fully automated graph theory research.}




\maketitle

\section{Introduction}

This paper introduces the \emph{Optimist}, an autonomous agent for generating conjectures in graph theory that iteratively adapts its output based on feedback. The \emph{Optimist} builds upon the principles of \emph{TxGraffiti}~\cite{TxGraffiti-I} with enhancements in mixed-integer programming, heuristic search, and a memory-based structure that allow it to efficiently produce and refine conjectures. Through its agent-based framework, the \emph{Optimist} can systematically generate inequalities involving graph invariants and filter results based on empirical strength, novelty, and known mathematical knowledge. The system, implemented in Python, is open-source and accompanied by Jupyter notebooks, enabling accessibility and reproducibility available at this papers companion GitHub repository\footnote{\url{https://github.com/RandyRDavila/The-Optimist/tree/main}}. 

A key feature of the \emph{Optimist} is its dynamic memory structure, which stores computed graph invariants and theorems, facilitating rapid retrieval and incremental updates as new graphs are introduced. This adaptive structure allows the system to refine its conjectures with minimal human intervention, moving toward a form of fully automated reasoning in graph theory. Additionally, the \emph{Optimist} is designed to interact with a complementary (human \emph{or machine}) agent, the \emph{Pessimist}, which will challenge the \emph{Optimist}’s conjectures by identifying counterexamples. Together, these agents form a dueling framework, termed \emph{GraphMind}, enabling continuous conjecture generation, verification, and refinement in a closed feedback loop.

The remainder of this paper is organized as follows: Section~\ref{sec:related-work} reviews prior work in automated conjecture generation, situating the \emph{Optimist} within this field. Section~\ref{sec:methods} details the system’s architecture, heuristics, and optimization methods. Section~\ref{sec:results} presents the conjectures generated by the \emph{Optimist}, highlighting both rediscovered known results and new inequalities. Finally, Section~\ref{sec:conclusion} discusses the future work and the development of the \emph{Optimist} in tandem with \emph{Pessimist} and the potential impact of \emph{GraphMind} in advancing automated reasoning in graph theory.

\section{Related Work}\label{sec:related-work}

The concept of intelligent machines contributing to mathematical research dates back to Turing’s 1948 proposal, which envisioned machines capable of high-level reasoning with minimal external data~\cite{Turing}. This vision set the stage for early computer-assisted mathematics. Among the first systems to embody this idea was Newell and Simon's \emph{Logic Theorist}, developed in the 1950s, which demonstrated the potential for machines to prove theorems in first-order logic. Newell and Simon anticipated that computers would one day play a central role in mathematical discovery~\cite{SimonNewell}. While early systems like the \emph{Logic Theorist} were focused on theorem proving, concurrent efforts in computer-assisted conjecturing emerged with Wang’s \emph{Program II}~\cite{Wang}, designed to generate statements interpretable as conjectures in logic. A significant challenge in Wang’s work, however, was filtering meaningful statements from the overwhelming volume of output. As Wang noted:
\begin{quote}
“The number of theorems printed out after running the machine for a few hours is so formidable that the writer has not even attempted to analyze the mass of data obtained”~\cite{Wang}.
\end{quote}

Refining large lists of generated statements into meaningful conjectures became a focal challenge in automated conjecture systems, prompting later programs to develop heuristics for identifying statements of mathematical significance.

\subsection{Fajtlowicz's \emph{Graffiti} and the Dalmatian Heuristic}\label{subsec:Graffiti}
A major advancement in filtering meaningful conjectures emerged with Fajtlowicz’s \emph{Graffiti} program in the 1980s. \emph{Graffiti} generated inequalities between real-valued functions (invariants) on mathematical objects, primarily within graph theory. The program operated on a database of graphs, aiming to identify inequalities of the form:
\[
\alpha \leq f(\text{invariants}),
\]
where $\alpha$ is a target invariant and $f(\text{invariants})$ is an expression involving other invariants, often in the form of sums, products, or more complex combinations constructed through an arithmetic expression tree. For example, \emph{Graffiti} might conjecture an inequality such as $\alpha \leq \frac{n}{2}$, where $\alpha$ represents the independence number of a graph, and $n$ denotes the graph's order (number of vertices).

The challenge of distinguishing nontrivial conjectures from trivial ones persisted. To address this, \emph{Graffiti} introduced the innovative \emph{Dalmatian} heuristic, a technique for refining conjectures based on their significance and validity across known examples. The Dalmatian heuristic operates with two primary criteria:

\noindent\textbf{The Dalmatian Heuristic}
\begin{itemize}
  \item \textbf{Truth Test:} The conjectured inequality must hold for all graphs in the database.
  \item \textbf{Significance Test:} The conjecture must provide a stronger bound for at least one graph compared to previously generated conjectures.
\end{itemize}

Fajtlowicz described the heuristic's process as follows:
\begin{quote}
``The program keeps track of conjectures made in the past, and when it encounters a new candidate for a conjecture, it first verifies whether an example in the database shows that the conjecture does not follow from earlier conjectures. If no such example exists, the conjecture is rejected as non-informative. If one exists, the program proceeds with testing the conjecture’s correctness, and then checks whether the conjecture should be rejected by other heuristics. If accepted, the list of conjectures is revised, and less informative conjectures are removed from the list and stored separately in case the new conjecture is refuted later.''~\cite{Fajtlowicz-V-1995}
\end{quote}

The Dalmatian heuristic not only reduced the number of conjectures but also enhanced their relevance by prioritizing conjectures that advanced current knowledge. When applied to identifying bounds on invariants, particularly those of active research interest, \emph{Graffiti} produced conjectures where existing theory offered insufficient predictions for invariant values. Such conjectures, if extending current understanding, were considered significant. Indeed, \emph{Graffiti} generated numerous conjectures that contributed substantially to both \emph{graph theory}~\cite{Fajtlowicz-Waller-1986, Chung-1988, Alon-Seymour-1989, Fajtlowicz-Clemson-1989, Beezer-1989, Favaron-1990, Favaron-Research-1991, Favaron-1991, Favaron-DM-1993, Griggs-Kleitman-1994, Fajtlowicz-McColgan-1995, Wang-1997, Firby-1997, Bollobas-Erdos-1998, Bollobas-Riordan-1998, Caro-1998, Dankelmann-Swart-Oellermann-1998, Jelen-1999, Codenotti-2000, Beezer-Riegsecker-Smith-2001, Favaron-2003, Zhang-2004, Dankelmann-Dlamini-Swart-2005, Hansen-2009, Cygan-2012, Yue-2019} and \emph{mathematical chemistry}~\cite{Fowler-1997, Fowler-1998, Fowler-1999, Fajtlowicz-Larson-2003, Stevanovic-Caporossi-2005, Fajtlowicz-Buckminsterfullerene-2005, Fajtlowicz-John-Sach-2005, Doslic-Reti-2011}. For further details on \emph{Graffiti} and the Dalmatian heuristic, see~\cite{TxGraffiti-I, Larson}.

\subsection{Extensions of the Dalmatian Heuristic: \emph{Graffiti.pc} and \emph{Conjecturing}}\label{subsec:Extensions}

Building upon the Dalmatian heuristic, DeLaViña’s \emph{Graffiti.pc} and Larson’s \emph{Conjecturing} programs extended automated conjecturing in graph theory and beyond. Developed during DeLaViña’s PhD studies under Fajtlowicz, \emph{Graffiti.pc} retained the original focus on graph theory while significantly expanding computational capabilities. Unlike its predecessor, which handled datasets with a few hundred graphs, \emph{Graffiti.pc} was implemented in C++ to work with much larger datasets, often comprising millions of graphs, and sometimes required computational clusters for efficient conjecture generation and testing. With applications in both education and research, \emph{Graffiti.pc} notably contributed to domination theory, leading to important results in this field~\cite{Henning-Yeo-2013, Henning-Yeo-2014, Henning-Wash-2017}. Despite these advances, increasing the dataset size did not always yield higher quality conjectures, as both \emph{Graffiti} and \emph{Graffiti.pc} were limited by the absence of certain small, special graphs in their datasets, which could serve as counterexamples to conjectures.

Recognizing the importance of selecting relevant datasets, Larson’s \emph{Conjecturing} program introduced a more flexible framework capable of generating conjectures across various mathematical disciplines, while still implementing the Dalamatian heuristic. Although the focus remained on inequalities involving graph invariants, \emph{Conjecturing} allowed for user-defined objects, invariants, and relations, expanding beyond graph theory into fields such as number theory and algebra. This adaptability enabled researchers to tailor the conjecture generation process for specific problems and customize parameters to guide the program towards nontrivial results.

\subsection{\emph{TxGraffiti}}\label{subsec:TxGraffiti}
\emph{TxGraffiti}~\cite{TxGraffiti-I} is an automated conjecturing program maintained by the author that focuses on generating meaningful conjectures in graph theory by leveraging principles similar to Fajtlowicz’s \emph{Graffiti}, particularly the emphasis on the mathematical strength of statements. However, \emph{TxGraffiti} applies linear optimization models to a precomputed tabular dataset of graph invariants, generating both upper and lower bounds for target invariants by solving optimization problems that minimize or maximize the discrepancy between a target invariant and an expression involving other invariants. Specifically, to establish an upper bound, \emph{TxGraffiti} minimizes a linear objective subject to constraints that enforce the inequality across all graphs meeting certain boolean conditions. This approach results in inequalities of the form $\alpha \leq m \cdot f(\text{invariants}) + b$, where $\alpha$ is the target invariant, $f(\text{invariants})$ represents other graph properties, and $m$ and $b$ are optimized scalars. Similarly, lower bounds are generated by maximizing this objective under reversed constraints, offering a dual approach to bounding invariants within the dataset.

To prioritize conjectures, \emph{TxGraffiti} employs the touch number heuristic, which measures how often an inequality holds as an equality across different graphs, thus reflecting the conjecture's strength and relevance. Conjectures with high touch numbers are prioritized, as they suggest a tighter relationship between invariants. Additionally, \emph{TxGraffiti} uses the static-Dalmatian heuristic, a refinement of the original Dalmatian heuristic, to eliminate redundant conjectures by filtering out those that can be derived from others via transitivity. This heuristic operates on a fixed set of conjectures, iteratively discarding any conjecture that does not introduce new information relative to previously accepted ones. Together, the optimization methods and heuristics used by \emph{TxGraffiti} have led to several published results in graph theory (see~\cite{TxGraffiti-2023, CaDaPe2020, CaDaHePe2022, DaHe21a, DaHe19b, DaHe19c}).

For other notable automated conjecturing systems that have contributed to various mathematical domains, see Lenat’s \emph{AM}~\cite{Lenat_1, Lenat_2, Lenat_3}, Epstein’s \emph{Graph Theorist}~\cite{Epstein_2}, Colton’s \emph{HR} system~\cite{Colton_1, Colton_2, Colton_3}, Hansen and Caporossi’s \emph{AutoGraphiX} (\emph{AGX})~\cite{AGX_1, AGX_2}, and Mélot’s \emph{GraPHedron} and its successor \emph{PHOEG}~\cite{graphedron_1, devillez2019}.

\section{The \emph{Optimist}: An Automated Conjecturing Agent}\label{sec:methods}

The \emph{Optimist} is an automated conjecturing agent dedicated to conjecture generation in graph theory, designed to construct, evaluate, and refine conjectures systematically. Drawing inspiration from the principles underlying \emph{TxGraffiti}, \emph{Optimist} implements a range of computational techniques aimed at producing conjectures with mathematical rigor and relevance. The \emph{Optimist} incorporates foundational components, including graph data, invariants, and a repository of known theorems, which serve as the basis for its conjecture generation.

\emph{Optimist} distinguishes itself from static conjecturing tools through its adaptability and agent-like autonomy. By employing iterative refinement and adaptive updates, \emph{Optimist} dynamically integrates new information, such as additional theorems or counterexamples, which allows it to iteratively enhance the quality and accuracy of its conjectures. This capability to evolve its knowledge base and refine its conjectures in response to new data characterizes \emph{Optimist} as a continuously learning system within automated reasoning, with the potential to contribute robustly to the development of theoretical insights in graph theory.

\subsection{Initial Knowledge and Setup}

At the core of the \emph{Optimist} agent’s framework is an initial set of structured inputs that constitute its foundational knowledge, forming the basis for conjecture formulation. This setup includes three primary components: a collection of graphs, a dictionary of computable graph invariant functions, and a repository of known theorems. Each graph in the collection provides a distinct context in which relationships between invariants can be examined, thereby supplying diverse examples to inform potential conjectures. 

The dictionary of invariants encompasses functions that compute essential properties of each graph, such as independence number, vertex count, and bipartiteness, which serve as key components in constructing conjectures. These invariants provide quantifiable data points that the \emph{Optimist} utilizes in identifying relationships and generating bounds.

Furthermore, \emph{Optimist} references a repository of known theorems to prevent the generation of redundant conjectures that merely replicate established results. This enables the system to focus on conjectures that extend beyond current knowledge, prioritizing relationships that are novel or have yet to be formally proven.

To facilitate conjecture generation and analysis, \emph{Optimist} employs several internal data structures. The framework organizes conjectures into separate lists of upper and lower bounds for each target invariant, consolidating these conjectures into a comprehensive collection. This approach allows for efficient retrieval and streamlined analysis, supporting a systematic examination of conjectures across various graph invariants.

\subsection{Constructing a Knowledge Base from Graph Invariants}

To systematically investigate relationships between graph invariants, the \emph{Optimist} framework constructs a tabular dataset, termed its \emph{knowledge base}, or \emph{knowledge table} (implemented as a \emph{Pandas DataFrame}). In this structure, each row corresponds to a unique graph instance, while each column represents a specific invariant or boolean property. This organization enables efficient data access and facilitates conjecture generation by allowing the framework to process invariant data in a manner akin to feature-based analysis in machine learning models. Analogous to how learning models identify patterns across features, \emph{Optimist} identifies potential conjectures by exploring relationships among these stored invariants.

A primary advantage of this knowledge base is that it stores all invariant values upon initial computation, obviating the need for recomputation during conjecture generation and validation. Given that calculating graph invariants can be computationally intensive, this storage approach significantly reduces the overhead of conjecture generation, streamlining the exploration process. By storing these invariant values centrally, \emph{Optimist} can efficiently retrieve relationships among invariants in real time, thereby facilitating more dynamic conjecture formulation.

This knowledge base thus functions as a structured “memory” for the \emph{Optimist} agent, enabling it to effectively leverage previously computed data in the same way a reasoning system might recall facts to inform subsequent decision-making. Through this design, \emph{Optimist} not only enhances computational efficiency but also supports a more comprehensive exploration of graph-theoretic relationships.

\subsection{Mixed-Integer Programming for Conjecture Generation}
\label{sec:mip-method}

The \emph{Optimist} framework employs a mixed-integer programming (MIP) approach to generate conjectures establishing upper and lower bounds on a target invariant. Unlike \emph{TxGraffiti}, which typically relates a single invariant to the target invariant, \emph{Optimist} formulates conjectures by considering linear combinations of multiple invariants under various boolean conditions. The \emph{Optimist} agent simultaneously seeks both upper and lower bounds, and in cases where bounds converge, it identifies exact equalities.

To produce meaningful bounds on a target invariant, \emph{Optimist} solves two MIP formulations that optimize for tight upper and lower bounds. The framework emphasizes maximizing instances where each bound holds as an equality, ensuring conjectures that are not only accurate but also sharp. This focus on equality instances aligns with the goal of producing conjectures that provide informative bounds on extreme values of the target invariant.

Let $\alpha$ denote the target invariant, and let $f(\text{invariants})$ represent a linear combination of other invariants. The bounds we seek can be expressed as:
\[
\alpha \leq f(\text{invariants}) + b \quad \text{(upper bound)} \quad \text{and} \quad \alpha \geq f(\text{invariants}) + b \quad \text{(lower bound)},
\]
where $f(\text{invariants})$ is a weighted sum of invariants with weights that maximize equality conditions across a dataset derived from the \emph{Optimist} agent’s knowledge base. Let $Y_i$ represent the target invariant for each graph $i$, and let $\mathbf{X}_i$ represent the selected invariants of graph $i$. The MIP seeks weights $w_1, w_2, \dots, w_k$ and an intercept $b$ to optimize the following constraints:

For the upper bound, we solve:
\[
Y_i \leq \sum_{j=1}^{k} w_j^{\text{upper}} X_{ij} + b^{\text{upper}}, \quad \text{for each graph } i.
\]
Binary variables $z_i^{\text{upper}}$ are introduced to indicate equality, expressed by the constraint:
\[
\left(\sum_{j=1}^{k} w_j^{\text{upper}} X_{ij} + b^{\text{upper}} \right) - Y_i \leq M (1 - z_i^{\text{upper}}),
\]
where $M$ is a large constant that relaxes the equality requirement when $z_i^{\text{upper}} = 0$.

Similarly, for the lower bound, we solve:
\[
Y_i \geq \sum_{j=1}^{k} w_j^{\text{lower}} X_{ij} + b^{\text{lower}}, \quad \text{for each graph } i.
\]
Binary variables $z_i^{\text{lower}}$ are introduced, with constraints defined as:
\[
Y_i - \left(\sum_{j=1}^{k} w_j^{\text{lower}} X_{ij} + b^{\text{lower}} \right) \leq M (1 - z_i^{\text{lower}}).
\]

The objective function maximizes the total number of equality instances:
\[
\max \sum_i z_i^{\text{upper}} + \sum_i z_i^{\text{lower}}.
\]
This formulation identifies optimal weights and intercepts for the bounds, yielding:
\[
\alpha \leq \sum_{j=1}^{k} w_j^{\text{upper}} X_j + b^{\text{upper}} \quad \text{and} \quad \alpha \geq \sum_{j=1}^{k} w_j^{\text{lower}} X_j + b^{\text{lower}}.
\]
When the solutions for upper and lower bounds converge, we obtain an equality:
\[
\alpha = \sum_{j=1}^{k} w_j^{\text{upper}} X_j + b^{\text{upper}} = \sum_{j=1}^{k} w_j^{\text{lower}} X_j + b^{\text{lower}}.
\]
Otherwise, the inequalities define a bounded range for $\alpha$, delineating the behavior of the target invariant over the graph dataset.

The MIP is implemented in Python using the \emph{PuLP} library, providing a flexible and efficient platform for solving these models; see Appendix~\ref{appendix:inequality-generation} or our GitHub repository\footnote{\url{https://github.com/RandyRDavila/The-Optimist/tree/main}}. Through this optimization approach, \emph{Optimist} generates conjectures that not only align with observed data but also enhance interpretability by prioritizing equality instances, thus identifying sharp bounds for the target invariant.

\subsection{Systematic Generation of Conjectures}

The \emph{Optimist} framework explores diverse combinations of invariants systematically, aiming to generate a comprehensive set of conjectures for each target invariant. For each target invariant, \emph{Optimist} constructs bounds by iterating through pairs of explanatory invariants, each conditioned by a single boolean property. This process is managed through the function \texttt{make\_all\_linear\_conjectures}, which organizes these combinations to generate upper and lower bounds; see Listing~\ref{lst:inequality-generation}.

The function \texttt{make\_all\_linear\_conjectures} performs the following operations:
\begin{itemize}
    \item For each pair of invariants, it applies all specified boolean properties, ensuring that the conjecture generation covers a broad space of possible relationships.
    \item Within each pair, the target invariant is constrained to avoid redundancies, ensuring that neither explanatory invariant directly corresponds to the target.
    \item The function calls the MIP-based \texttt{make\_linear\_conjectures} function to compute bounds on the target invariant, yielding a candidate set of upper and lower conjectures.
\end{itemize}

The output of this procedure is a set of candidate conjectures for each target invariant, reflecting a systematic approach to exploring multi-invariant relationships under varied boolean conditions.

\begin{lstlisting}[style=pythonstyle, caption={Generating All Possible Linear Bounds on a Target Invariant}, label={lst:inequality-generation}]
def make_all_linear_conjectures(df, target, others, properties):
    # Create conjectures for every pair of invariants in 'others' combined with each property.
    upper_conjectures = []
    lower_conjectures = []
    seen_pairs = []
    for other1, other2 in combinations(others, 2):
        set_pair = set([other1, other2])
        if set_pair not in seen_pairs:
            seen_pairs.append(set_pair)
            for prop in properties:
                # Ensure that neither of the 'other' invariants equals the target.
                if other1 != target and other2 != target:
                    # Generate the conjecture for this combination of two invariants - solve an MIP model.
                    upper_conj, lower_conj = make_linear_conjectures(df, target, [other1, other2], hyp=prop)
                    upper_conjectures.append(upper_conj)
                    if lower_conj:
                        lower_conjectures.append(lower_conj)

    return upper_conjectures, lower_conjectures
\end{lstlisting}

This function thus enables \emph{Optimist} to generate bounds across a systematic range of invariant combinations, producing a foundational set of conjectures that can then be refined and prioritized based on empirical support and mathematical relevance.

\subsection{The Hazel Heuristic}
\label{sec:hazel-heuristic}

After invoking the \texttt{make\_all\_linear\_conjectures} function on a target invariant, the \emph{Optimist} agent typically produces extensive lists of potential conjectures—often in the hundreds—depending on the numerical and boolean properties considered. To identify and prioritize conjectures with strong empirical backing, \emph{Optimist} applies the \emph{Hazel Heuristic}, a filtering process centered on each conjecture’s \emph{touch number}.

The touch number of a conjectured inequality is defined as the number of graphs in the \emph{Optimist} knowledge base for which the inequality holds as an equality. A high touch number suggests that the conjectured bound is not only generally valid but also sharp for a significant subset of graphs, indicating a potentially fundamental relationship between the target invariant and the explanatory invariants. High-touch conjectures are, therefore, more likely to capture structural properties of graph invariants and yield insights beyond loose or incidental bounds.

The \emph{Hazel Heuristic} applies three successive steps:
\begin{enumerate}
    \item \textbf{Deduplication}: Removes duplicate conjectures to ensure the uniqueness of bounds.
    \item \textbf{Filtering}: Discards conjectures with touch numbers below a specified threshold, removing bounds that lack consistent sharpness across the dataset.
    \item \textbf{Sorting}: Orders conjectures in descending order of touch number, bringing conjectures with the highest empirical support to the forefront.
\end{enumerate}

\begin{lstlisting}[style=pythonstyle, caption={Hazel Heuristic Implementation}, label={lst:hazel}]
def hazel_heuristic(conjectures, min_touch=0):
    # Remove duplicate conjectures.
    conjectures = list(set(conjectures))

    # Remove conjectures never attaining equality.
    conjectures = [conj for conj in conjectures if conj.touch>min_touch]

    # Sort the conjectures by touch number.
    conjectures.sort(key=lambda x: x.touch, reverse=True)

    # Return the sorted list of conjectures.
    return conjectures
\end{lstlisting}

By prioritizing conjectures with the highest touch numbers, the \emph{Hazel Heuristic} ensures that the most empirically significant conjectures are advanced for further analysis. This ranking approach confers several advantages:
\begin{itemize}
    \item \textbf{Relevance}: Conjectures with high touch numbers are more likely to represent tight bounds, making them strong candidates for formal validation or proof.
    \item \textbf{Robustness}: A high touch number indicates that a conjecture’s bound is consistently representative across diverse graph structures, enhancing its robustness and potential generalizability.
    \item \textbf{Informational Value}: High-touch conjectures highlight areas where graph invariants exhibit strong interactions, offering insights into fundamental invariant relationships.
\end{itemize}

Through the \emph{Hazel Heuristic}, \emph{Optimist} focuses on conjectures with maximal empirical evidence, refining the conjecture set to emphasize inequalities with the greatest potential for meaningful mathematical contribution.

\subsection{The \emph{Morgan} Heuristic}
\label{sec:morgan-heuristic}

The \emph{Morgan} Heuristic identifies and removes redundant conjectures, prioritizing the most general conjectures with broad applicability. Specifically, a conjecture is flagged as redundant if it shares the same conclusion as another but is based on a more specific (less general) hypothesis. By focusing on generality, this heuristic ensures that the final set of conjectures includes only the most powerful and widely applicable statements.

For example, consider the two conjectures in Listing~\ref{lst:morgan_example}. Every tree is a bipartite graph, but not every bipartite graph is a tree. Therefore, Conjecture 4, which applies to all connected bipartite graphs, is more general than Conjecture 2, which applies only to trees. As such, Conjecture 2 is discarded in favor of Conjecture 4.

\begin{lstlisting}[style=conjecturestyle, caption={Two example conjectures on $\alpha$.}, label={lst:morgan_example}]
Conjecture 2. If G is a tree, then independence_number = n - matching_number. This bound holds with equality on 3 graphs.

Conjecture 4. If G is a connected and bipartite graph, then independence_number = n - matching_number. This bound holds with equality on 5 graphs.
\end{lstlisting}

In practice, the generality of a conjecture’s hypothesis is determined by counting the number of graphs in the \emph{Optimist} knowledge base that satisfy the hypothesis. If one hypothesis applies to a larger subset of graphs than another, it is considered more general. This count-based approach, though heuristic, provides a practical method for determining generality even when relationships between graph classes are not fully defined; see the implementation in Listing~\ref{lst:morgan}. 

\begin{lstlisting}[style=pythonstyle, caption={Morgan Heuristic}, label={lst:morgan}]
def morgan_heuristic(conjectures):
    new_conjectures = conjectures.copy()
    for conj_one in conjectures:
        for conj_two in new_conjectures.copy():  # Make a copy for safe removal
            # Avoid comparing the conjecture with itself
            if conj_one != conj_two:
                # Check if conclusions are the same and conj_one's hypothesis is more general
                if conj_one.conclusion == conj_two.conclusion and conj_one.hypothesis > conj_two.hypothesis:
                    new_conjectures.remove(conj_two)  # Remove the less general conjecture (conj_two)

    return new_conjectures
\end{lstlisting}

The key steps in the Morgan Heuristic are as follows:

\begin{enumerate}
    \item \textbf{Identify Redundancy}: Conjectures are flagged as redundant if they share identical conclusions but differ in hypothesis generality.
    \item \textbf{Assess Generality}: For each pair of conjectures with the same conclusion, the conjecture with a hypothesis that applies to a larger subset of graphs is considered more general, based on empirical counts from the \emph{Optimist} dataset.
    \item \textbf{Comparison and Removal}: If one conjecture is a generalization of another, the less general conjecture is removed.
    \item \textbf{Refinement}: This process iteratively refines the conjecture set, ensuring that only unique, maximally general conjectures remain.
\end{enumerate}

Applying the \emph{Morgan} Heuristic yields a refined list of conjectures that are maximally general and devoid of redundancy. This reduction not only simplifies the conjecture set but also enhances its theoretical strength, highlighting statements with the broadest potential applicability.

\subsection{The \emph{Strong-Smokey} and \emph{Weak-Smokey} Heuristics}
\label{sec:smokey-heuristics}

To further refine the conjecture set, the \emph{Optimist} agent employs two selective heuristics: the \emph{weak-smokey} and \emph{strong-smokey} heuristics. Both heuristics reduce redundancy by examining the set of \emph{sharp graphs}—instances where each conjectured inequality holds as an equality—though each applies distinct criteria to retain only conjectures that offer unique insights.

The \emph{weak-smokey} heuristic iterates through conjectures in descending order of touch number, retaining conjectures that introduce new sharp graphs not already covered by previously selected conjectures. This approach prioritizes conjectures with complementary equality instances, achieving a balance between inclusiveness and informativeness. By focusing on unique sharp instances, \emph{weak-smokey} substantially reduces the initial conjecture set while maintaining broad empirical relevance. 
\begin{lstlisting}[style=pythonstyle, caption={Weak-Smokey Heuristic}, label={lst:weak-smokey}]
def weak_smokey(conjectures):
    # Start with the conjecture that has the highest touch number.
    conj = conjectures[0]

    # Initialize the list of strong conjectures.
    strong_conjectures = [conj]

    # Get the set of sharp graphs.
    sharp_graphs = conj.sharps

    # Iterate over the remaining conjectures in the list.
    for conj in conjectures[1:]:
        if conj.is_equal():
            # Save all equality conjectures.
            strong_conjectures.append(conj)
            sharp_graphs = sharp_graphs.union(conj.sharps)
        else:
            # Check if the current conjecture shares the same sharp graphs as any already selected strong conjecture.
            if any(conj.sharps.issuperset(known.sharps) for known in strong_conjectures):
                # If it does, add the current conjecture to the list of strong conjectures.
                strong_conjectures.append(conj)
                # Update the set of sharp graphs to include the newly discovered sharp graphs.
                sharp_graphs = sharp_graphs.union(conj.sharps)
            # Otherwise, check if the current conjecture introduces new sharp graphs (graphs where the conjecture holds).
            elif conj.sharps - sharp_graphs != set():
                # If new sharp graphs are found, add the conjecture to the list.
                strong_conjectures.append(conj)
                # Update the set of sharp graphs to include the newly discovered sharp graphs.
                sharp_graphs = sharp_graphs.union(conj.sharps)

    # Return the list of strong, non-redundant conjectures.
    return strong_conjectures
\end{lstlisting}

In contrast, the \emph{strong-smokey} heuristic applies a stricter criterion, requiring that each retained conjecture covers a strict superset of sharp graphs compared to any previously selected conjecture. This process begins with the conjecture having the highest touch number, then adds conjectures only if they introduce strictly new equality instances. By requiring that each new conjecture expands the set of sharp graphs, \emph{strong-smokey} yields a minimal set of conjectures, often halving the number retained by \emph{weak-smokey}. This minimality ensures that only conjectures providing the most unique insights are retained.
\begin{lstlisting}[style=pythonstyle, caption={Strong-Smokey Heuristic}, label={lst:strong-smokey}]
def strong_smokey(conjectures):
    # Start with the conjecture that has the highest touch number.
    conj = conjectures[0]

    # Initialize the list of strong conjectures.
    strong_conjectures = [conj]

    # Get the set of sharp graphs.
    sharp_graphs = conj.sharps

    # Iterate over the remaining conjectures in the list.
    for conj in conjectures[1:]:
        if conj.is_equal():
            # Save all equality conjectures.
            strong_conjectures.append(conj)
        else:
            # Check if the current conjecture set of sharp graphs is a superset of any already selected strong conjecture.
            if any(conj.sharps.issuperset(known.sharps) for known in strong_conjectures):
                # If it does, add the current conjecture to the list of strong conjectures.
                strong_conjectures.append(conj)
                sharp_graphs = sharp_graphs.union(conj.sharps)

    # Return the list of strong, non-redundant conjectures.
    return strong_conjectures
\end{lstlisting}

Applying either the \emph{weak-smokey} or \emph{strong-smokey} heuristic significantly reduces the conjecture set for a given target invariant, typically retaining only four or five conjectures from an initially large list. As a stricter filter, \emph{strong-smokey} produces a subset of the conjectures selected by \emph{weak-smokey}. It is employed when \emph{Optimist} seeks to retain only the most general and minimal set of conjectures, emphasizing statements that provide the greatest informational value.

\subsection{Conjecture Generation, Filtering, and Adaptive Knowledge Update}
\label{sec:conjecture-generation}

The \emph{Optimist} agent is designed as an iterative system for conjecture generation, filtering, and continuous refinement based on new information. At the core of this process is the use of mixed-integer programming (MIP) to generate conjectures. Using combinations of graph invariants, \emph{Optimist} generates potential upper and lower bounds for a target invariant. Each conjecture is initially stored in memory, organized by target invariant.

Given the potentially large number of conjectures generated, \emph{Optimist} applies a multi-stage filtering process to ensure that only the most informative and robust conjectures are retained. The first phase employs the \emph{Hazel} heuristic, which prioritizes conjectures with the highest empirical relevance by sorting them according to \emph{touch number}—a measure of how frequently each conjecture holds as an equality across the dataset. This step ensures that \emph{Optimist} focuses on conjectures that provide tight bounds for the target invariant.

Next, the \emph{Morgan} heuristic is applied to remove redundant conjectures with more restrictive hypotheses that do not contribute new insights over more general conjectures with the same conclusion. This heuristic ensures the final conjecture set retains only the most widely applicable statements.

A final layer of filtering applies either the \emph{strong-smokey} or \emph{weak-smokey} heuristic, depending on the desired strictness. The \emph{weak-smokey} heuristic favors inclusiveness, retaining conjectures that add new sharp instances, while the \emph{strong-smokey} heuristic strictly selects conjectures that cover strictly broader sets of sharp graphs. This stage produces a minimal, highly informative set of conjectures, prioritizing those with the broadest empirical applicability -- essentially equivalent to an unpublished version of \emph{TxGraffiti} called \emph{TxGraffiti {II}}, which is available as an interactive website\footnote{\url{https://txgraffiti.streamlit.app/Generate_MIP_Conjectures}}.

\begin{figure}[ht]
    \centering
    \includegraphics[width=0.85\textwidth]{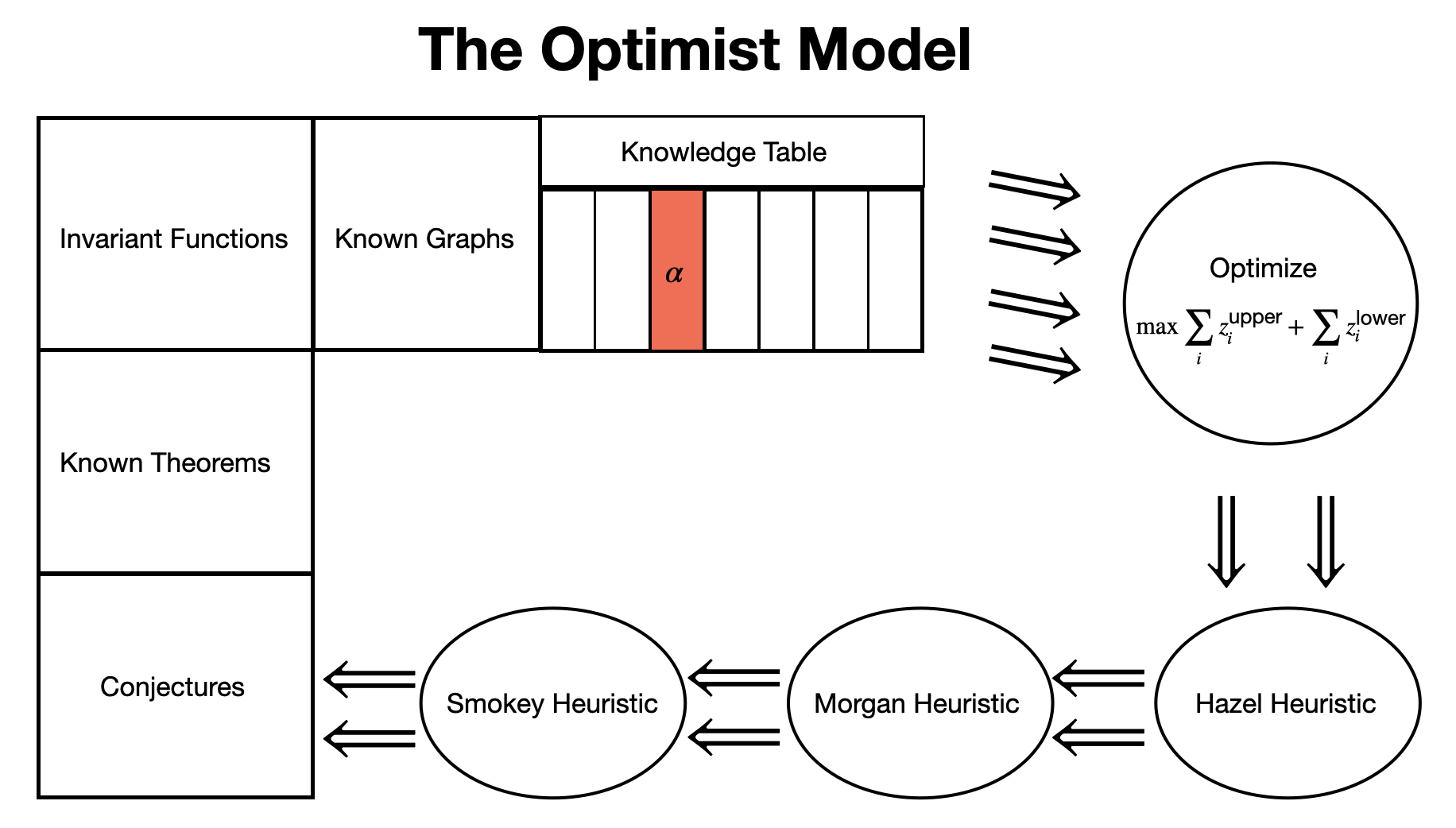} 
    \caption{A diagram illustrating the \emph{Optimist} agent's process for generating, filtering, and updating conjectures when targeting the invariant $\alpha$.}
    \label{fig:example_image}
\end{figure}

\paragraph{Knowledge Update and Counterexample Handling} \label{sec:knowledge-update}
The \emph{Optimist} agent is designed to adapt dynamically to new data, allowing it to refine its conjectures and avoid rediscovering known results. Before retaining any conjecture, \emph{Optimist} references a repository of established theorems, removing any conjecture implied by these known results. This approach focuses the agent’s attention on novel insights.

The agent further adapts based on counterexamples provided by users. If a counterexample graph contradicts an existing conjecture, \emph{Optimist} incorporates the new graph into its knowledge base and updates all conjectures to ensure consistency with observed data. This self-improvement process enables the agent to maintain accurate and reliable conjectures, continuously refining its knowledge and conjecture quality with each iteration. By iterating through these stages of generation, filtering, and adaptive updates, \emph{Optimist} exemplifies a robust approach to autonomous reasoning and knowledge refinement in graph theory.

\section{Case Study: Conjecture Generation for the Independence Number}\label{sec:results}

This section presents a case study to evaluate the \emph{Optimist} agent’s ability to generate conjectures that are both empirically grounded and theoretically insightful. Our objective is to demonstrate the \emph{Optimist}’s effectiveness in autonomously rediscovering classical bounds on the independence number, $\alpha(G)$, and its adaptability in refining conjectures in response to iterative user feedback.

\emph{Optimist} was assessed on its capacity to conjecture bounds on $\alpha(G)$, a fundamental but computationally challenging graph invariant. Beginning with a minimal dataset, we examined whether the agent could autonomously derive meaningful bounds on $\alpha(G)$ and subsequently refine its conjectures through feedback and new information.

\subsection{Experimental Setup and Initial Conjecture Generation}
\begin{figure}[ht]
    \centering
    \subfloat[]{\includegraphics[width=0.3\textwidth]{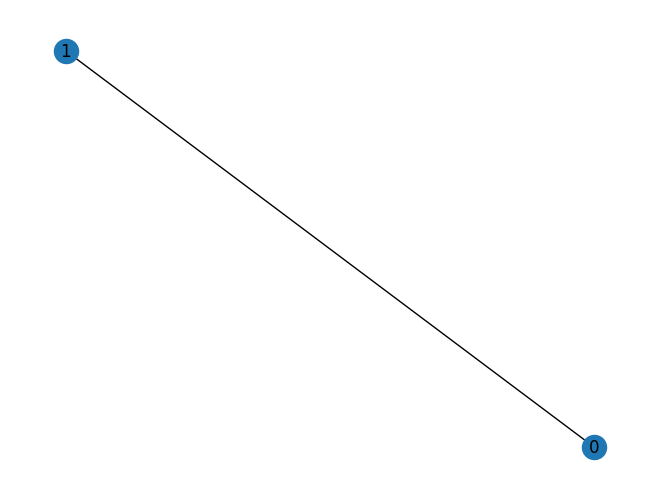}} \hfill
    \subfloat[]{\includegraphics[width=0.3\textwidth]{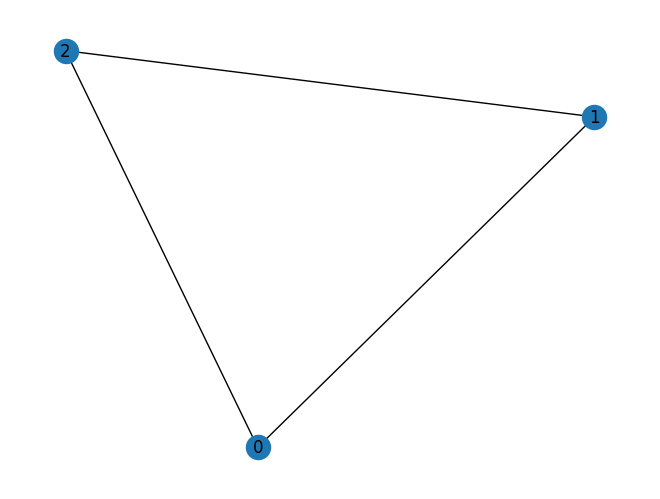}} \hfill
    \subfloat[]{\includegraphics[width=0.3\textwidth]{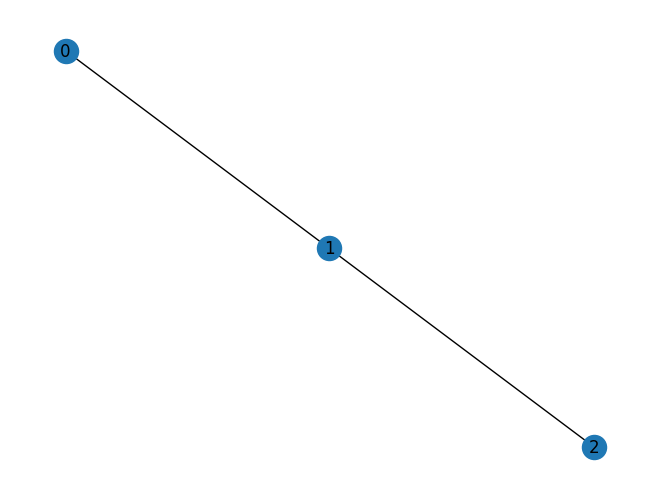}}
    \caption{The initial set of graphs in \emph{Optimist}’s knowledge base. (a) The complete graph $K_2$ (also the path $P_2$), (b) the complete graph $K_3$ (also the cycle graph $C_3$), (c) The path graph $P_3$.}
    \label{fig:graphs}
\end{figure}

The initial setup for \emph{Optimist} included three of the smallest nontrivial connected graph structures: the complete graphs $K_2$ and $K_3$ and the path graph $P_3$, as shown in Figure~\ref{fig:graphs}. The agent was configured with a selection of invariants relevant to conjecturing $\alpha(G)$, including the order $n$, minimum degree $\delta$, maximum degree $\Delta$, matching number $\mu$, and minimum maximal matching number $\mu^*$, among others. With this setup, \emph{Optimist} could explore a variety of relationships between $\alpha(G)$ and these properties.

From this minimal configuration, \emph{Optimist} generated an initial series of conjectures aimed at establishing upper and lower bounds for $\alpha(G)$. The resulting conjectures were broad, reflecting the limited dataset and covering a variety of possible inequalities. This initial series included nearly 60 conjectures, illustrating the heuristic’s dependency on dataset size and the need for further refinement to isolate the most significant bounds.

\subsection{Iterative Knowledge Refinement through User Feedback}

Following the initial conjecture set, the user identified conjectures that could be invalidated by counterexamples. For instance, the conjecture presented in Listing~\ref{lst:false1} was shown to be false for the path graph $P_6$.
\begin{lstlisting}[style=conjecturestyle, caption={An early false conjecture on the independence number $\alpha$.}, label={lst:false1}]
Conjecture 1. If G is a connected graph, then independence_number = order - minimum_degree
\end{lstlisting}
Noting this counterexample, the user informed the \emph{Optimist} agent of $P_6$, prompting the agent to update its knowledge base and regenerate its conjecture set.
\begin{lstlisting}[style=pythonstyle, caption={Informing the Agent}, label={lst:update1}]
optimist.update_graph_knowledge(nx.path_graph(6))
\end{lstlisting}

\begin{figure}[ht]
    \centering
    \subfloat[]{\includegraphics[width=0.3\textwidth]{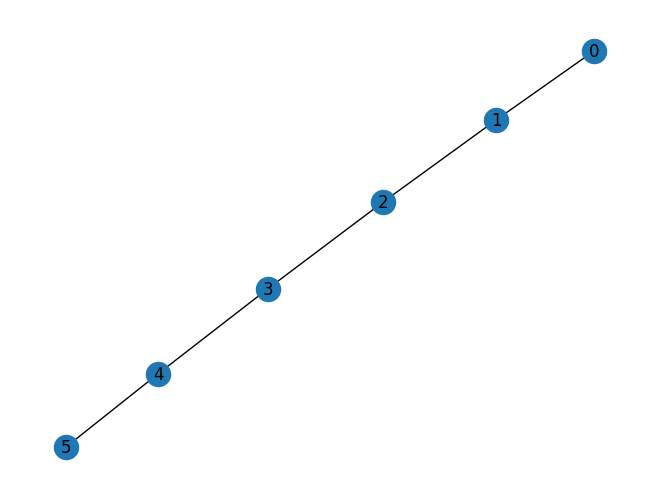}} \hfill
    \subfloat[]{\includegraphics[width=0.3\textwidth]{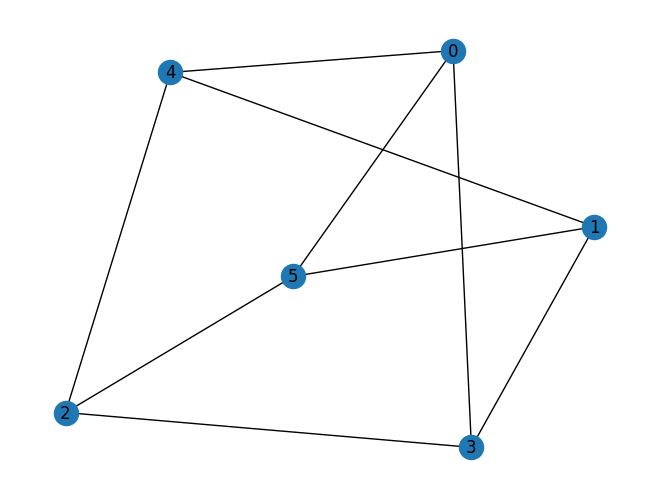}} \hfill
    \subfloat[]{\includegraphics[width=0.3\textwidth]{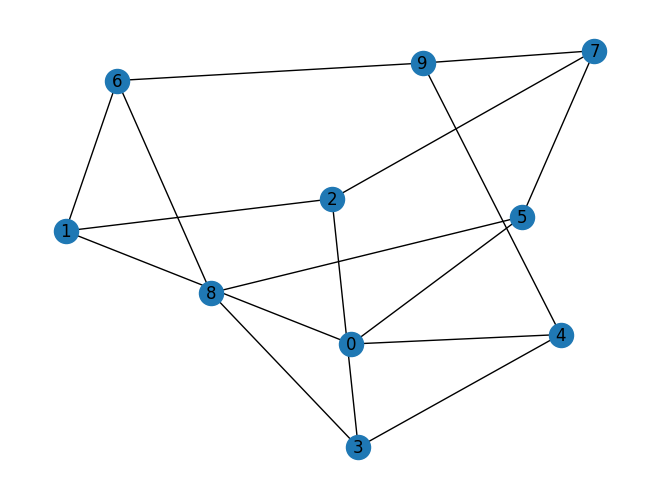}} \\
    \subfloat[]{\includegraphics[width=0.3\textwidth]{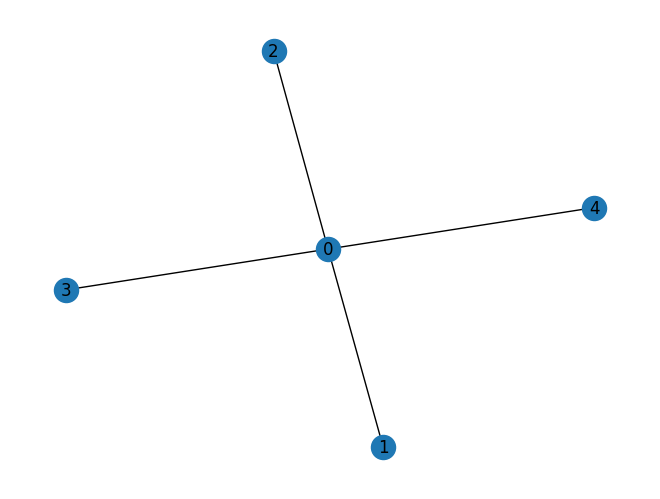}} \hfill
    \subfloat[]{\includegraphics[width=0.3\textwidth]{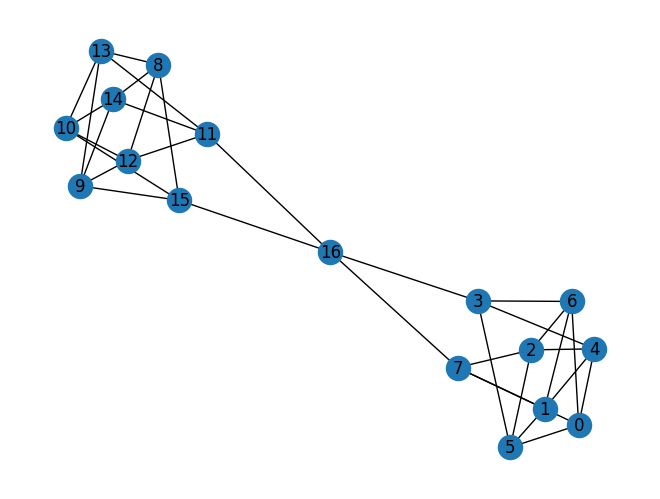}} \hfill
    \subfloat[]{\includegraphics[width=0.3\textwidth]{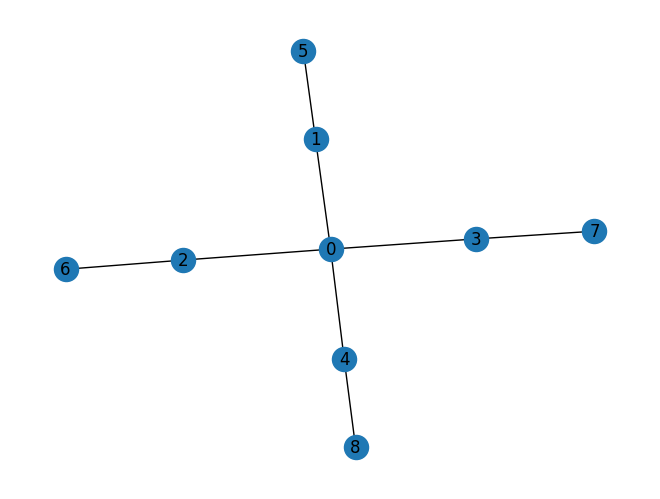}} \\
    \subfloat[]{\includegraphics[width=0.3\textwidth]{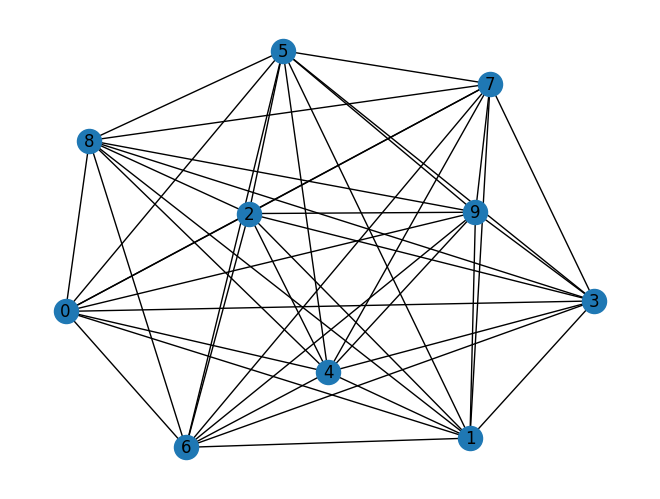}} \hfill
    \subfloat[]{\includegraphics[width=0.3\textwidth]{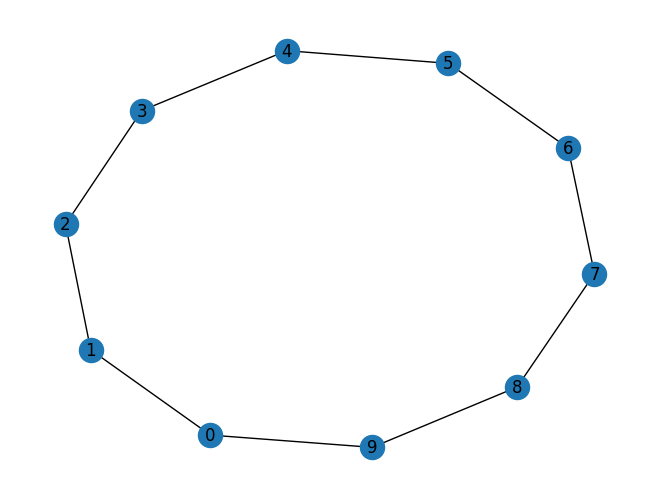}} \hfill
    \subfloat[]{\includegraphics[width=0.3\textwidth]{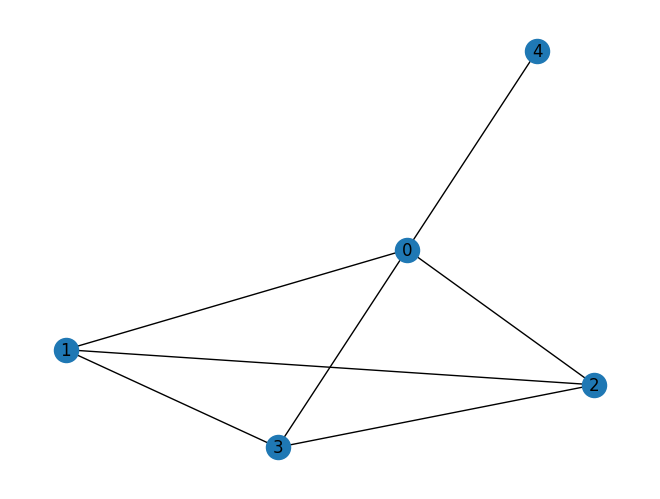}}
    \caption{The set of counterexamples introduced to the \emph{Optimist} agent during the iterative feedback process of working with a human \emph{Pessimist} -- notably some of the counterexamples being nontrivial (see graphs (c) and (e))}
    \label{fig:image_grid}
\end{figure}

With each new graph added, \emph{Optimist} refined its conjecture set, retaining inequalities that consistently held across the expanded dataset and discarding or modifying those that failed to generalize. This interactive process continued over multiple iterations, with \emph{Optimist} generating conjectures and the user acting as a \emph{Pessimist} by providing counterexamples. Each update expanded the agent’s knowledge base, enabling it to generate more reliable conjectures. Figure~\ref{fig:image_grid} shows the set of counterexample graphs introduced during this process.

This iterative refinement process allowed \emph{Optimist} to progressively narrow down the set of plausible bounds on $\alpha(G)$, adapting its conjectures in response to new data. The user’s counterexamples acted as a form of hypothesis testing, whereby each graph introduced was a probe challenging \emph{Optimist}’s emerging hypotheses. Through this feedback loop, the agent evolved toward a more reliable and theoretically grounded set of conjectures.

Occasionally, the \emph{Optimist} agent generated conjectures that aligned with known bounds or equations for the independence number, often appearing at the top of its conjecture lists. When these familiar relationships were recognized by the user, they were communicated to the agent, enabling \emph{Optimist} to store this knowledge and exclude known results from its conjecture set going forward. This process allowed \emph{Optimist} to focus on unexplored relationships. See Listing~\ref{lst:update-known-theorems} for an example of updating the agent with known results.

\begin{lstlisting}[style=pythonstyle, caption={Informing the Agent of Known Theorems}, label={lst:update-known-theorems}]
known_theorems = [optimist.upper_conjectures["independence_number"][0],]
optimist.update_known_theorems(known_theorems)
\end{lstlisting}

By the conclusion of the experiment, after providing nine counterexamples, the \emph{Optimist} agent had learned several key theorems, including well-known and classical results in graph theory. These learned theorems are shown in Listing~\ref{lst:known-theorems}.

\begin{lstlisting}[style=conjecturestyle, caption={The Agent's Learned Theorems}, label={lst:known-theorems}]
Theorem. If G is a connected graph, then independence_number <= order - minimum_degree

Theorem. If G is a connected graph, then independence_number <= order - matching_number

Theorem. If G is a connected and bipartite graph, then independence_number = order - matching_number

Theorem. If G is a connected and bipartite graph, then independence_number >= maximum_degree

Theorem. If G is a connected and regular graph, then independence_number <= matching_number

Theorem. If G is a connected and bipartite graph, then independence_number >= 1/2 * order
\end{lstlisting}

The emergence of these theorems with minimal user intervention highlights \emph{Optimist}’s empirical robustness and adaptability, demonstrating its potential to align autonomously with established mathematical knowledge. This adaptive capacity reflects the system’s strength in identifying both classical and novel bounds, reinforcing its value as a tool for automated reasoning in graph theory. Please see our GitHub repository\footnote{\url{https://github.com/RandyRDavila/The-Optimist/tree/main}} for the complete experiment available as a Colab and Jupyter notebook. 

\subsection{The Evolution of the \emph{Optimist} Agent and the Concept of GraphMind}

The \emph{Optimist} agent begins as a knowledge-based system, equipped with a foundational set of graphs and a suite of functions for computing graph invariants. From this initial setup, the agent constructs a tabular dataset, representing each graph as a unique instance and each function as a feature. This tabular structure serves as a dynamic memory, enabling \emph{Optimist} to generate conjectures based on observed relationships across invariants, initially producing bounds and identities for specific graph properties.

\begin{figure}[ht]
    \centering
    \begin{tikzpicture}[node distance=5.0cm, auto, thick] 
        \node (optimist) [rectangle, draw, rounded corners, fill=blue!10, text width=3cm, align=center] {\textbf{Optimist Agent} \\ (Conjecture Generation)};
        \node (conjecture) [circle, draw, fill=gray!10, text width=2cm, align=center, right of=optimist] {\textbf{Conjecture}};
        \node (pessimist) [rectangle, draw, rounded corners, fill=red!10, text width=3cm, align=center, right of=conjecture, node distance=5cm] {\textbf{Pessimist Agent} \\ (Counterexample Search)};
        \node (counterexample) [circle, draw, fill=gray!10, text width=3.2cm, align=center, below of=pessimist, node distance=4cm] {\textbf{Counterexample} \\ (A Graph Instance)};
        \node (knowledge) [rectangle, draw, rounded corners, fill=green!10, text width=3cm, align=center, below of=optimist, node distance=4cm] {\textbf{Knowledge Base}};

        \draw[->] (optimist) -- (conjecture) node[midway, above] {Generates};
        \draw[->] (conjecture) -- (pessimist) node[midway, above] {Presents};
        \draw[->] (pessimist) -- (counterexample) node[midway, right] {Finds};
        \draw[->] (counterexample) -- (knowledge) node[midway, below] {Adds to};
        \draw[->] (knowledge.north) -- (optimist.south) node[midway, left] {Refines}; 
        
        \draw[->, dashed] (knowledge.north) -| (optimist); 

    \end{tikzpicture}
    \caption{The feedback loop within the \textbf{GraphMind} framework, illustrating the iterative interaction between the \emph{Optimist} (conjecture generation) and \emph{Pessimist} (counterexample search) agents, and the refinement of the knowledge base.}
    \label{fig:feedback_loop}
\end{figure}
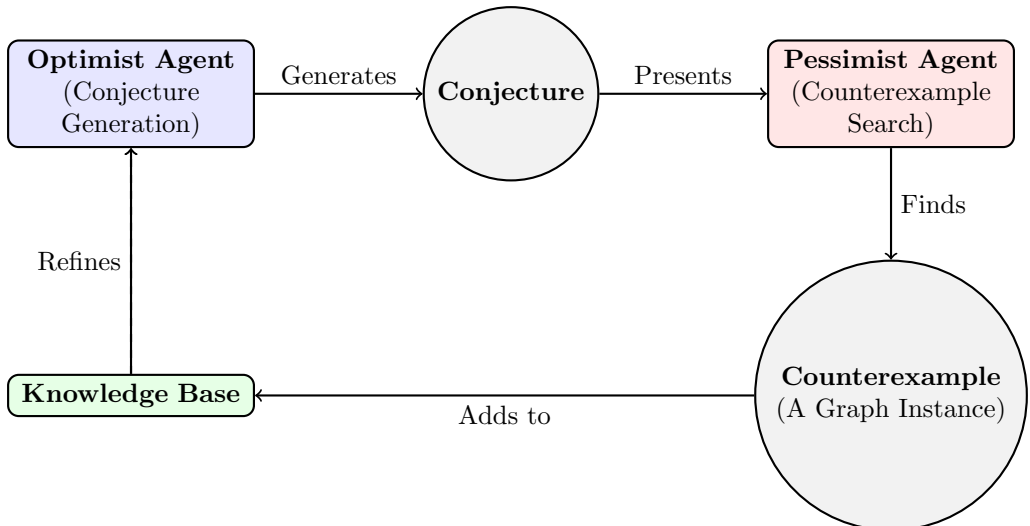

Through interaction with a counterexample generator—referred to as the \emph{Pessimist}, which can be either a human user or an autonomous agent—\emph{Optimist} adapts its conjectures and knowledge base. When a counterexample is found that disproves a conjecture, the \emph{Pessimist} introduces a new graph, prompting \emph{Optimist} to re-evaluate its existing conjectures. This iterative feedback loop allows \emph{Optimist} to refine its dataset, accumulate new graph structures, and distill empirically valid conjectures. Over time, \emph{Optimist} builds a repository of both new conjectures and established theorems, increasing its capability to generate stronger, more general conjectures that remain consistent with an ever-growing dataset; effectively increasing its intelligence; see Figure~\ref{fig:feedback_loop} for an illustration. 

In a future implementation where the \emph{Pessimist} is also an autonomous agent, this feedback loop could operate without human intervention. The \emph{Optimist} would continuously propose conjectures, while the \emph{Pessimist} systematically seeks counterexamples. This interaction forms an automated cycle of discovery, where conjecture generation and counterexample search evolve concurrently, refining both conjectures and the underlying graph structures in the dataset. This setup, while unable to formally prove conjectures, would create a self-sustaining environment for empirical discovery in graph theory—a system we refer to as \textbf{GraphMind}.

GraphMind represents a conceptual leap toward a fully automated graph theorist. While it lacks formal proof capabilities, its ability to autonomously discover conjectures and graph structures could drive meaningful advances in graph theory. By automating the processes of hypothesis generation and empirical verification, GraphMind would serve as a powerful tool for mathematical exploration, potentially uncovering new classes of graphs, invariants, and relationships that extend beyond human intuition.

\section{Conclusion and Future Directions}\label{sec:conclusion}

In this paper, we presented the \emph{Optimist} agent, a framework designed for automated conjecture generation in graph theory. Through the use of mixed-integer programming, heuristic filtering, and iterative refinement, \emph{Optimist} has demonstrated the ability to autonomously derive empirically strong and mathematically relevant conjectures. In particular, our case study on the independence number $\alpha(G)$ highlighted \emph{Optimist}’s capability to rediscover established results, such as Kőnig’s theorem and classical bounds, while dynamically refining its conjectures based on iterative user feedback. Notably, the \emph{Optimist} agent often conjectures published conjectures of \emph{TxGraffiti} (see~\cite{TxGraffiti-2023, CaDaPe2020, CaDaHePe2022, DaHe21a, DaHe19b, DaHe19c}) as a subset of other conjectures which remain open and under investigation. 

The \emph{Optimist} framework represents an important step toward an automated approach to hypothesis generation in discrete mathematics. By integrating a counterexample-searching counterpart—the \emph{Pessimist}—we envision a collaborative system, \textbf{GraphMind}, where \emph{Optimist} proposes conjectures and \emph{Pessimist} seeks counterexamples. This dueling-agent framework creates a feedback loop, allowing both agents to iteratively test, refine, and adapt conjectures without human intervention.

GraphMind represents a vision of a fully automated graph theorist. While current implementations do not have formal proof capabilities, the ability to generate, validate, and refine conjectures could lead to significant advancements in empirical discovery within graph theory. As both agents operate in tandem, they provide a balanced and rigorous exploration of graph invariants, potentially uncovering novel relationships and classes of graphs that might remain hidden to human intuition.

Future work will focus on developing an autonomous \emph{Pessimist} agent to enable a continuous cycle of conjecture and counterexample generation. Early results suggest that deep reinforcement learning, combined with Monte Carlo Tree Search, can empower the \emph{Pessimist} agent to construct counterexample graphs on a small number of vertices. Additionally, integrating formal proof techniques, such as automated theorem proving, could further enhance GraphMind, allowing it not only to propose conjectures but also to rigorously verify or disprove them. Furthermore, leveraging language models to identify and cross-reference known theorems in the literature could effectively eliminate the need for human intervention in GraphMind. By fully automating this exploratory process, GraphMind could serve as a powerful tool for mathematical discovery, providing researchers with a novel and efficient approach to exploring complex structures in graph theory.

\backmatter

\begin{appendices}

\section{Python Code for Inequality Generation}
\label{appendix:inequality-generation}

\begin{lstlisting}[style=pythonstyle, caption={Generating Inequalities with PuLP}]
def make_linear_conjectures(df, target, others, hyp="a connected graph", symbol="G",):
    """
    Generates upper and lower bound multilinear conjectures for a target variable
    using a mixed-integer programming approach to maximize the number of equalities at extreme values.

    Parameters
    ----------
    df : pandas.DataFrame
        The DataFrame containing graph data and associated variables (invariants).
    target : str
        The name of the target variable (dependent variable) in the conjecture.
    others : list of str
        A list of other variable names (independent variables or invariants) to be used in the conjecture.
    hyp : str, optional
        The name of the hypothesis variable, indicating the condition applied to graphs (default is "a connected graph").
    symbol : str, optional
        The symbol representing the object in the conjecture (default is "G").

    Returns
    -------
    tuple
        A tuple containing:
        - MultiLinearConjecture for the upper bound.
        - MultiLinearConjecture for the lower bound (if different slopes exist), else None.
    """
    pulp.LpSolverDefault.msg = 0

    # Filter data for the hypothesis condition.
    df = df[df[hyp] == True]
    true_objects = df["name"].tolist()

    # Preprocess the data to find the maximum Y for each X for the upper bound
    df_upper = df.loc[df.groupby(others)[target].idxmax()]
    # Preprocess the data to find the minimum Y for each X for the lower bound
    df_lower = df.loc[df.groupby(others)[target].idxmin()]

    # Extract the data for the upper and lower bound problems
    Xs_upper = [df_upper[other].tolist() for other in others]
    Y_upper = df_upper[target].tolist()
    Xs_lower = [df_lower[other].tolist() for other in others]
    Y_lower = df_lower[target].tolist()

    # Initialize the MIP problem.
    prob = pulp.LpProblem("Maximize_Equality", pulp.LpMaximize)

    # Initialize the variables for the MIP (one set for upper bound and one for lower bound).
    ws_upper = [pulp.LpVariable(f"w_upper{i+1}", upBound=4, lowBound=-4) for i in range(len(others))]  # Weights for upper bound
    ws_lower = [pulp.LpVariable(f"w_lower{i+1}", upBound=4, lowBound=-4) for i in range(len(others))]  # Weights for lower bound
    b_upper = pulp.LpVariable("b_upper", upBound=3, lowBound=-3)
    b_lower = pulp.LpVariable("b_lower", upBound=3, lowBound=-3)

    # Binary variables z_j^upper and z_j^lower to maximize equality conditions for extreme points
    z_upper = [pulp.LpVariable(f"z_upper{j}", cat="Binary") for j in range(len(Y_upper))]
    z_lower = [pulp.LpVariable(f"z_lower{j}", cat="Binary") for j in range(len(Y_lower))]

    M = 1000  # Big-M value

    # Upper bound constraints (maximize equality on max Y values)
    for j in range(len(Y_upper)):
        prob += pulp.lpSum([ws_upper[i] * Xs_upper[i][j] for i in range(len(others))]) + b_upper >= Y_upper[j]
        prob += pulp.lpSum([ws_upper[i] * Xs_upper[i][j] for i in range(len(others))]) >= b_upper
        prob += pulp.lpSum([ws_upper[i] * Xs_upper[i][j] for i in range(len(others))]) + b_upper - Y_upper[j] <= M * (1 - z_upper[j])

    # Lower bound constraints (maximize equality on min Y values)
    for j in range(len(Y_lower)):
        prob += pulp.lpSum([ws_lower[i] * Xs_lower[i][j] for i in range(len(others))]) + b_lower <= Y_lower[j]
        prob += pulp.lpSum([ws_lower[i] * Xs_lower[i][j] for i in range(len(others))]) >= b_lower
        prob += Y_lower[j] - pulp.lpSum([ws_lower[i] * Xs_lower[i][j] for i in range(len(others))]) - b_lower <= M * (1 - z_lower[j])

    # Maximize the number of equalities for both upper and lower bounds
    prob += pulp.lpSum(z_upper) + pulp.lpSum(z_lower)

    # Solve the MIP
    prob.solve()

    if prob.status != 1:
        print("No feasible solution found.")
        return None
    else:
        weights_upper = [Fraction(w.varValue).limit_denominator(10) for w in ws_upper]
        weights_lower = [Fraction(w.varValue).limit_denominator(10) for w in ws_lower]
        b_upper_value = Fraction(b_upper.varValue).limit_denominator(10)
        b_lower_value = Fraction(b_lower.varValue).limit_denominator(10)

        if weights_lower == weights_upper and b_upper_value == b_lower_value:
            touch_upper = len(true_objects)

            # Create the hypothesis and conclusion objects for both upper and lower bounds.
            hypothesis = Hypothesis(hyp, true_object_set=true_objects)
            upper_conclusion = MultiLinearConclusion(target, "=", weights_upper, others, b_upper_value)

            # Return the full conjecture object (not the conclusion directly).
            return MultiLinearConjecture(hypothesis, upper_conclusion, symbol, touch_upper, true_objects), None
        else:
            Xs_true_upper = [df[other].tolist() for other in others]
            Y_true_upper = df[target].tolist()
            Xs_true_lower = [df[other].tolist() for other in others]
            Y_true_lower = df[target].tolist()
            # Compute the number of instances of equality - the touch number of the conjecture.
            touch_set_upper = set([true_objects[j] for j in range(len(Y_true_upper)) if
                                Y_true_upper[j] == sum(weights_upper[i] * Xs_true_upper[i][j] for i in range(len(others))) + b_upper_value])
            touch_set_lower = set([true_objects[j] for j in range(len(Y_true_lower)) if Y_true_lower[j] == sum(weights_lower[i] * Xs_true_lower[i][j] for i in range(len(others))) + b_lower_value])

            touch_upper = len(touch_set_upper)
            touch_lower = len(touch_set_lower)

            # Create the hypothesis and conclusion objects for both upper and lower bounds.
            hypothesis = Hypothesis(hyp, true_object_set=true_objects)
            upper_conclusion = MultiLinearConclusion(target, "<=", weights_upper, others, b_upper_value)
            lower_conclusion = MultiLinearConclusion(target, ">=", weights_lower, others, b_lower_value)

            # Return the full conjecture object (not the conclusion directly).
            return MultiLinearConjecture(hypothesis, upper_conclusion, symbol, touch_upper, touch_set_upper), \
                MultiLinearConjecture(hypothesis, lower_conclusion, symbol, touch_lower, touch_set_lower)

def make_all_linear_conjectures(df, target, others, properties):
    """
    Generates upper and lower bound multilinear conjectures for every unique pair of invariants in 'others'
    combined with each property in 'properties'. Calls `make_linear_conjectures` for each combination.

    Parameters
    ----------
    df : pandas.DataFrame
        The DataFrame containing graph data and associated variables (invariants).
    target : str
        The name of the target variable in the conjecture.
    others : list of str
        A list of other variable names (independent variables or invariants).
    properties : list of str
        A list of properties (hypotheses) to apply for conjecture generation.

    Returns
    -------
    tuple
        A tuple containing:
        - A list of upper bound MultiLinearConjecture objects for each valid pair of variables and property.
        - A list of lower bound MultiLinearConjecture objects (if they exist) for each valid pair of variables and property.
    """
    upper_conjectures = []
    lower_conjectures = []
    seen_pairs = []
    for other1, other2 in combinations(others, 2):
        set_pair = set([other1, other2])
        if set_pair not in seen_pairs:
            seen_pairs.append(set_pair)
            for prop in properties:
                # Ensure that neither of the 'other' invariants equals the target.
                if other1 != target and other2 != target:
                    # Generate the conjecture for this combination of two invariants.
                    upper_conj, lower_conj = make_linear_conjectures(df, target, [other1, other2], hyp=prop)
                    upper_conjectures.append(upper_conj)
                    if lower_conj:
                        lower_conjectures.append(lower_conj)

    return upper_conjectures, lower_conjectures

\end{lstlisting}

\section{Python Code Implementing Heuristics}
\label{appendix:heuristics}

\begin{lstlisting}[style=pythonstyle, caption={Optimist Heuristics}]
def hazel_heuristic(conjectures, min_touch=0):
    """
    Filters and sorts a list of conjectures based on touch number.

    This heuristic:
    - Removes duplicate conjectures.
    - Removes conjectures that never attain equality (touch <= min_touch).
    - Sorts the remaining conjectures in descending order of touch number.

    Parameters
    ----------
    conjectures : list of Conjecture
        The list of conjectures to filter and sort.
    min_touch : int, optional
        The minimum touch number required for a conjecture to be retained (default is 0).

    Returns
    -------
    list of Conjecture
        The sorted list of conjectures with the highest touch numbers.
    """
    # Remove duplicate conjectures.
    conjectures = list(set(conjectures))

    # Remove conjectures never attaining equality.
    conjectures = [conj for conj in conjectures if conj.touch > min_touch]

    # Sort the conjectures by touch number.
    conjectures.sort(key=lambda x: x.touch, reverse=True)

    # Return the sorted list of conjectures.
    return conjectures


def morgan_heuristic(conjectures):
    """
    Removes redundant conjectures based on generality.

    A conjecture is considered redundant if another conjecture has the same conclusion
    and a more general hypothesis (i.e., its true_object_set is a superset of the redundant one).

    Parameters
    ----------
    conjectures : list of Conjecture
        The list of conjectures to filter.

    Returns
    -------
    list of Conjecture
        A list with redundant conjectures removed.
    """
    new_conjectures = conjectures.copy()

    for conj_one in conjectures:
        for conj_two in new_conjectures.copy():  # Make a copy for safe removal
            # Avoid comparing the conjecture with itself
            if conj_one != conj_two:
                # Check if conclusions are the same and conj_one's hypothesis is more general
                if conj_one.conclusion == conj_two.conclusion and conj_one.hypothesis > conj_two.hypothesis:
                    new_conjectures.remove(conj_two)  # Remove the less general conjecture (conj_two)

    return new_conjectures


def weak_smokey(conjectures):
    """
    Selects conjectures based on equality and distinct sharp graphs.

    This heuristic:
    - Starts with the conjecture having the highest touch number.
    - Retains conjectures that either satisfy equality or introduce new sharp graphs.

    Parameters
    ----------
    conjectures : list of Conjecture
        The list of conjectures to filter.

    Returns
    -------
    list of Conjecture
        A list of strong conjectures with distinct or new sharp graphs.
    """
    # Start with the conjecture that has the highest touch number (first in the list).
    conj = conjectures[0]

    # Initialize the list of strong conjectures with the first conjecture.
    strong_conjectures = [conj]

    # Get the set of sharp graphs (i.e., graphs where the conjecture holds as equality) for the first conjecture.
    sharp_graphs = conj.sharps

    # Iterate over the remaining conjectures in the list.
    for conj in conjectures[1:]:
        if conj.is_equal():
            strong_conjectures.append(conj)
            sharp_graphs = sharp_graphs.union(conj.sharps)
        else:
            # Check if the current conjecture shares the same sharp graphs as any already selected strong conjecture.
            if any(conj.sharps.issuperset(known.sharps) for known in strong_conjectures):
                # If it does, add the current conjecture to the list of strong conjectures.
                strong_conjectures.append(conj)
                # Update the set of sharp graphs to include the newly discovered sharp graphs.
                sharp_graphs = sharp_graphs.union(conj.sharps)
            # Otherwise, check if the current conjecture introduces new sharp graphs (graphs where the conjecture holds).
            elif conj.sharps - sharp_graphs != set():
                # If new sharp graphs are found, add the conjecture to the list.
                strong_conjectures.append(conj)
                # Update the set of sharp graphs to include the newly discovered sharp graphs.
                sharp_graphs = sharp_graphs.union(conj.sharps)

    # Return the list of strong, non-redundant conjectures.
    return strong_conjectures


def strong_smokey(conjectures):
    """
    Selects conjectures that strongly subsume others based on sharp graphs.

    This heuristic:
    - Starts with the conjecture having the highest touch number.
    - Retains conjectures whose sharp graphs are supersets of previously selected conjectures.

    Parameters
    ----------
    conjectures : list of Conjecture
        The list of conjectures to filter.

    Returns
    -------
    list of Conjecture
        A list of conjectures with non-redundant, strongly subsuming sharp graphs.
    """
    # Start with the conjecture that has the highest touch number (first in the list).
    conj = conjectures[0]

    # Initialize the list of strong conjectures with the first conjecture.
    strong_conjectures = [conj]

    # Get the set of sharp graphs (i.e., graphs where the conjecture holds as equality) for the first conjecture.
    sharp_graphs = conj.sharps

    # Iterate over the remaining conjectures in the list.
    for conj in conjectures[1:]:
        if conj.is_equal():
            strong_conjectures.append(conj)
        else:
            # Check if the current conjecture set of sharp graphs is a superset of any already selected strong conjecture.
            if any(conj.sharps.issuperset(known.sharps) for known in strong_conjectures):
                # If it does, add the current conjecture to the list of strong conjectures.
                strong_conjectures.append(conj)
                sharp_graphs = sharp_graphs.union(conj.sharps)

    # Return the list of strong, non-redundant conjectures.
    return strong_conjectures


def filter_false_conjectures(conjectures, df):
    """
    Filters conjectures to remove those with counterexamples in the provided data.

    Parameters
    ----------
    conjectures : list of Conjecture
        The list of conjectures to filter.
    df : pandas.DataFrame
        The DataFrame containing graph data.

    Returns
    -------
    list of Conjecture
        A list of conjectures with no counterexamples in the DataFrame.
    """
    new_conjectures = []
    for conj in conjectures:
        if conj.false_graphs(df).empty:
            new_conjectures.append(conj)
    return new_conjectures


def filter_false_conjectures(conjectures, df):
    new_conjectures = []
    for conj in conjectures:
        if conj.false_graphs(df).empty:
            new_conjectures.append(conj)
    return new_conjectures
\end{lstlisting}

\section{Python Code for the \texttt{Optimist} Python class}
\label{appendix:optimist}

\begin{lstlisting}[style=pythonstyle, caption={The \texttt{Optimist} Python class.}, label={lst:optimist}]
class Optimist:
    """
    An agent that generates, refines, and stores mathematical conjectures on graph properties.

    Attributes
    ----------
    graphs : list
        A list of initial graphs used for conjecture generation.
    invariants : dict
        A dictionary where keys are invariant names and values are functions that compute the invariant for a graph.
    known_theorems : list, optional
        A list of known theorems to filter redundant conjectures.

    Methods
    -------
    build_knowledge()
        Computes invariant values for each graph and stores them in a DataFrame.
    conjecture(target, use_strong_smokey=False)
        Generates conjectures for a target invariant using heuristics, filtering known theorems, and storing results.
    write_on_the_wall(target)
        Displays stored conjectures for a given target along with details on equality instances.
    update_graph_knowledge(graph)
        Adds a new graph to the knowledge base and updates the stored invariant values.
    update_known_theorems(theorems)
        Adds a list of theorems to the known theorems list to avoid redundant conjectures.
    filter_known_upper_bounds(target, index)
        Filters a specific upper bound by adding it to known theorems and re-generating conjectures.
    filter_known_lower_bounds(target, index)
        Filters a specific lower bound by adding it to known theorems and re-generating conjectures.
    """

    def __init__(self, graphs: list, invariants: dict, known_theorems=[]):
        """
        Initializes the Optimist agent.

        Parameters
        ----------
        graphs : list
            A list of initial graphs to start with.
        invariants : dict
            A dictionary where keys are function names and values are the functions themselves.
        known_theorems : list, optional
            A list of known theorems to filter out redundant conjectures.
        """
        self.graphs = graphs
        self.invariants = invariants
        self.upper_conjectures = {}
        self.lower_conjectures = {}
        self.all_conjectures = {}
        self.known_theorems = known_theorems
        self.use_strong_smokey=False

    def build_knowledge(self):
        """
        Computes invariant values for each graph and stores them in a DataFrame.

        Iterates over the list of graphs, applies invariant functions to each graph,
        and stores the results in a DataFrame for conjecture generation.
        """
        rows = []
        for i, G in enumerate(self.graphs):
            row = {"name": f"G{i}"}
            for name, function in self.invariants.items():
                row[name] = function(G)
            rows.append(row)
        self.df = pd.DataFrame(rows)

    def conjecture(self, target):
        """
        Generates and filters conjectures for a target invariant.

        Uses upper and lower bound heuristics to create conjectures for the target invariant.
        Applies filtering heuristics to remove false, redundant, and less significant conjectures.

        Parameters
        ----------
        target : str
            The target variable for conjecture generation.
        use_strong_smokey : bool, optional
            Whether to use the strong version of the Smokey heuristic (default is False).

        Returns
        -------
        tuple
            A tuple containing lists of upper and lower bound conjectures.
        """
        numerical_columns = self.df.select_dtypes(include=['int64', 'float64']).columns.tolist()
        boolean_columns = self.df.select_dtypes(include=['bool']).columns.tolist()
        upper_conjectures, lower_conjectures = make_all_linear_conjectures(self.df, target, numerical_columns, boolean_columns)
        upper_conjectures = filter_false_conjectures(upper_conjectures, self.df)
        lower_conjectures = filter_false_conjectures(lower_conjectures, self.df)
        upper_conjectures, lower_conjectures = hazel_heuristic(upper_conjectures, min_touch=1), hazel_heuristic(lower_conjectures, min_touch=1)
        upper_conjectures, lower_conjectures = morgan_heuristic(upper_conjectures), morgan_heuristic(lower_conjectures)
        if self.use_strong_smokey:
            upper_conjectures, lower_conjectures = strong_smokey(upper_conjectures), strong_smokey(lower_conjectures)
        else:
            upper_conjectures, lower_conjectures = weak_smokey(upper_conjectures), weak_smokey(lower_conjectures)
        self.upper_conjectures[target] = [conj for conj in upper_conjectures if conj not in self.known_theorems]
        self.lower_conjectures[target] = [conj for conj in lower_conjectures if conj not in self.known_theorems]
        self.all_conjectures[target] = upper_conjectures + lower_conjectures
        return upper_conjectures, lower_conjectures

    def write_on_the_wall(self, target):
        """
        Displays stored conjectures for a given target.

        Prints each conjecture along with details about the number of graphs
        where equality holds (touch number).

        Parameters
        ----------
        target : str
            The target variable whose conjectures are to be displayed.
        """
        if target not in self.upper_conjectures or target not in self.lower_conjectures:
            self.conjecture(target)
        for i, upper_conj in enumerate(self.upper_conjectures[target]):
            print(f"Conjecture {i}. {upper_conj}")
            print(f"With equality on {upper_conj.touch} graphs.\n")

        print()
        for i, lower_conj in enumerate(self.lower_conjectures[target]):
            print(f"Conjecture {i}. {lower_conj}")
            print(f"With equality on {lower_conj.touch} graphs.\n")

    def update_graph_knowledge(self, graph):
        """
        Adds a new graph to the agent's knowledge base and updates the DataFrame.

        Computes invariant values for the new graph and appends them to the DataFrame.

        Parameters
        ----------
        graph : networkx.Graph
            The new graph to be added.
        """
        self.graphs.append(graph)
        new_row = {"name": f"G{len(self.graphs)}"}
        for name, function in self.invariants.items():
            new_row[name] = function(graph)
        self.df = pd.concat([self.df, pd.DataFrame([new_row])], ignore_index=True)

    def update_known_theorems(self, theorems):
        """
        Adds new theorems to the list of known theorems to filter redundant conjectures.

        Parameters
        ----------
        theorems : list
            A list of theorems to add to the known theorems list.
        """
        for thm in theorems:
            self.known_theorems.append(thm)

    def filter_known_upper_bounds(self, target, index):
        """
        Filters a specific upper bound by adding it to known theorems and re-generating conjectures.

        Parameters
        ----------
        target : str
            The target variable whose upper bounds are being filtered.
        index : int
            The index of the conjecture to add to known theorems.
        """
        theorem = self.upper_conjectures[target][index - 1]
        self.known_theorems.append(theorem)
        self.conjecture(target)

    def filter_known_lower_bounds(self, target, index):
        """
        Filters a specific lower bound by adding it to known theorems and re-generating conjectures.

        Parameters
        ----------
        target : str
            The target variable whose lower bounds are being filtered.
        index : int
            The index of the conjecture to add to known theorems.
        """
        theorem = self.lower_conjectures[target][index - 1]
        self.known_theorems.append(theorem)
        self.conjecture(target)

\end{lstlisting}

\end{appendices}


\bibliography{sn-bibliography}

\end{document}